\pdfoutput=1

\documentclass[11pt]{article}

\usepackage[]{emnlp2021}

\usepackage{times}
\usepackage{latexsym}

\usepackage[T1]{fontenc}

\usepackage[utf8]{inputenc}

\usepackage{microtype}

\usepackage{hhline}
\usepackage{tablefootnote}
\usepackage{adjustbox}
\usepackage{multirow}
\usepackage{arydshln} 

\title{Robustness and Sensitivity of BERT Models\\Predicting Alzheimer's Disease from Text}

\author{Jekaterina Novikova \\
  Winterlight Labs / Toronto, Canada \\
  \texttt{jekaterina@winterlightlabs.com} \\}

\begin{document}
\maketitle
\begin{abstract}
Understanding robustness and sensitivity of BERT models predicting Alzheimer's disease from text is important for both developing better classification models and for understanding their capabilities and limitations. In this paper, we analyze how a controlled amount of desired and undesired text alterations impacts performance of BERT. We show that BERT is robust to natural linguistic variations in text. On the other hand, we show that BERT is not sensitive to removing clinically important information from text.
\end{abstract}

\section{Introduction}

Alzheimer's disease (AD) is a prevalent neurodegerative condition that inhibits cognitive abilities and impacts one’s language abilities. For example, cognitively impaired people tend to use more pronouns instead of nouns, and pause more often between sentences in narrative speech~\cite{roark2011spoken}. This insight makes automatic detection possible. Machine learning (ML) classifiers can detect cognitive impairments given descriptive linguistic features or using pre-trained large language models~\cite{balagopalan2018effect,zhu2019detecting}.

BERT is a model that achieves promising performance on a variety of tasks, including AD prediction from speech and language~\cite{Searle2020,Yuan2020}.
However, this promising performance may be fallacious, i.e. deep neural language models may learn pseudo patterns from training data to attain high performance on test sets~\citep{goyal2019making,gururangan-etal-2018-annotation,glockner-etal-2018-breaking,tsuchiya-2018-performance,geva-etal-2019-modeling}. Therefore,  in order to be confident in the outcomes of BERT models classifying AD it is important to assess whether these models are robust to some natural noise that may be introduced in language. It is also important to know if BERT models are sensitive to the aspects that are considered to be important for recognizing cognitive impairment from human language. 
 
 In this paper, we analyze robustness and sensitivity of BERT models in their ability to classify AD from text by analysing the effect of noise, introduced from artificial text perturbations, on the performance of the model. Some previous research was conducted on the impact of ASR-related noise on dementia detection~\cite{balagopalan2020impact}, as well as the effect of artificial text alterations on AD classification~\cite{novikova2019lexical}. However, these previous studies only focus on conventional classification models, such as Random Forest and SVM. To the best of our knowledge, we are the first to analyse how the noise introduced by texts perturbations impact BERT models, in the domain of AD classification.

\section{Methodology}

\subsection{Data}
\label{sec:dataset}

We use the ADReSS Challenge dataset~\cite{luz2020alzheimer}, which consists of 156 speech samples and associated transcripts from non-AD ($N$=78) and AD ($N$=78) English-speaking participants. Speech is elicited from participants through the Cookie Theft picture from the Boston Diagnostic Aphasia exam~\cite{goodglass2001bdae}.
In contrast to other datasets for AD detection such as DementiaBank's English Pitt Corpus~\cite{becker1994natural}, the ADReSS challenge dataset is well balanced in terms of age and gender (Table~\ref{tab:ds_compare}). 
Another benefit of this dataset is its division into standard train and test sets that makes it easy to directly compare to the previous research in the area. 

\begin{table}[t]
\begin{center}
\caption{Basic characteristics of the patients in each group in the ADReSS challenge dataset.}

\begin{adjustbox}{max width=0.8\linewidth}
\begin{tabular}{l|cl|c|c}
 \textbf{Dataset} & &  & \multicolumn{2}{c}{\textbf{Class}} \\
 & &  & \textbf{AD} & \textbf{Non-AD}\\ \hhline{=|==|=|=}
\multirow{2}{*}{ADReSS} & \multirow{2}{*}{Train} & Male & 24 & 24 \\

& & Female & 30 &  30 \\\cline{1-5}
\multirow{2}{*}{ADReSS} & \multirow{2}{*}{Test} & Male & 11 &  11 \\
 & & Female & 13 & 13  \\
\end{tabular}
\end{adjustbox}
\label{tab:ds_compare}
\end{center}
\end{table}

\subsection{Model} 
\label{sec:model}

Multiple recent studies showed that BERT is a promising model achieving strong enough performance in detecting Alzheimer's disease from transcribed speech~\cite{Searle2020,Yuan2020,Balagopalan2020,balagopalan2021comparing}. Motivated by these results, we use a fine-tuned BERT~\cite{devlin2019bert} model in this work. To leverage the language information encoded by BERT
~\cite{devlin2019bert}, 
we add a linear layer mapping representations from the final layer of a pre-trained 12-layer BERT base\footnote{https://huggingface.co/bert-base-uncased} for the AD vs non-AD binary classification task.  The transcript-level input to the  model consists of transcribed utterances with corresponding start and separator special tokens for each utterance, following Liu \emph{et al.}~\cite{liu2019text}. A pooled embedding
summarizing information across all tokens in the transcript is used as the aggregate transcript representation, and passed to the classification layer~\cite{devlin2019bert,wolf2019huggingface}. 
This model is then fine-tuned on training data for AD detection. 
For hyperparameter tuning, we optimize the number of epochs to 10 by varying it from 1 to 12 during cross-validation. 
Adam optimizer~\cite{kingma2014adam} and warmup linear learning rate scheduling~\cite{paszke2019pytorch} are used, based on prior work on fine-tuning BERT ~\cite{devlin2019bert,wolf2019huggingface}.

\subsection{Perturbation Approaches}
\label{sec:alterations}

We used a variety of word-based augmentation approaches with the help of the nlpaug\footnote{https://github.com/makcedward/nlpaug , the Python library for generating synthetic textual and speech data.} library to generate perturbed versions of the test set of the ADReSS dataset for the experiments. 

\textit{Back-translation: }this augmentation technique proposed by~\citet{sennrich-etal-2016-improving} leverages two translation models, one translating the source text from English to German and the other translating it back to English. Back-translated texts should maintain the semantics and basic syntactic structure of original texts and as such, robust model's performance should not decrease because of this augmentation.

\textit{Word substitution with synonyms: }following~\citet{niu2018adversarial}, we substitute a controlled varying amount of words in the transcript (10-90\%) with their synonyms in order to maintain semantic meaning of the utterances. Synonyms are extracted from the NLTK WordNet corpus\footnote{https://www.nltk.org/howto/wordnet.html}. Replacing words with their synonyms should not affect the ability of a robust model to accurately distinguish between healthy and AD classes.

\textit{Embedding-based word substitution: }following~\citet{alzantot-etal-2018-generating,wang-yang-2015-thats}, we use pre-trained word2vec embeddings to perform a KNN with cosine similarity search to find the similar word for replacement. We then substitute a varying subsets (from 10 to 90\%) of the original transcripts with these replacements. We hypothesize that model performance can be affected by such augmentation stronger than by synonym replacement, although this effect should not be significant for a robust AD prediction model.

\textit{Removal of filled pauses: }we remove all the filled pauses (transcribed as \textit{um} and \textit{uh}) from the original texts. Previous literature highlights the importance of pauses in Alzheimer's disease detection from speech~\cite{calley2010subjective, mack2013word, seifart2018nouns}. Several authors report increases in AD detection performance by extracting acoustic features such as filled pause counts~\cite{eyre-etal-2020-fantastic,toth2015automatic,toth2018speech,pistono2016pauses}. Removal of such information should make it more difficult for a model to accurately detect AD-related samples of text.

\textit{Removal of information units: } multiple studies of AD narratives in picture description tasks have reported the importance of information units in detecting cognitive impairment~\cite{fraser2016linguistic,croisile1996comparative}. Following \cite{croisile1996comparative}, we define four key categories of information units - subjects, locations, objects, and actions - and delete them from the original transcripts to generate perturbed versions of the test set. Such a removal should make it more difficult for a model to distinguish between healthy and AD samples. 

\section{Results}

The results of testing the fine-tuned BERT model on the variety of perturbed versions of the ADReSS test set show that the performance changes differently depending on different types of text alterations (Table~\ref{tab:perf}). Removing tokens of filled pauses does not change the performance at all. Removing information units, however, decreases the accuracy of the model by 4-8\%, depending on the type of the information unit. 
Back translation and substitutions of words with their synonyms or otherwise similar words also negatively affect performance of the model, although have the opposite effect on recall vs specificity. 

\begin{table}[t!]
\begin{adjustbox}{max width=1\linewidth, center}
\begin{tabular}{llllllll}
\begin{tabular}[c]{@{}l@{}}\textbf{Type of}\\\textbf{perturbation}\end{tabular} & \begin{tabular}[c]{@{}l@{}}\textbf{Level of}\\\textbf{perturbation}\end{tabular} & \textbf{Acc} & \textbf{F1} & \textbf{Prec} & \textbf{Rec} & \textbf{Spec} & \textbf{$W_1$}\\
\hline \hline
\begin{tabular}[c]{@{}l@{}}Original\\ transcript\end{tabular} & NA & 0.83 & 0.83 & 0.86 & 0.79 & 0.88 & NA\\
\hline
\begin{tabular}[c]{@{}l@{}}Deleting\\ filled pauses\end{tabular} & All & 0.83 & 0.83 & 0.86 & 0.79 & 0.88 & 2.40\\
\hdashline
\multirow{5}{*}{\begin{tabular}[c]{@{}l@{}}Deleting\\ information\\ units\end{tabular}} & All & 0.81 & 0.84 & 0.74 & 0.96 & 0.67 & 2.87\\
 & Action & 0.77 & 0.80 & 0.71 & 0.92 & 0.63 & 2.87\\
 & Location & 0.77 & 0.78 & 0.74 & 0.83 & 0.71 & 0.77\\
 & Object & 0.75 & 0.79 & 0.69 & 0.92 & 0.58 & 2.45\\
 & Subject & 0.79 & 0.79 & 0.79 & 0.79 & 0.79 & 2.13\\
 \hline
\begin{tabular}[c]{@{}l@{}}Back \\ translation\end{tabular} & Eng \textless-\textgreater DE & 0.75 & 0.75 & 0.75 & 0.75 & 0.75 & 6.02\\
 \hdashline
 \multirow{9}{*}{\begin{tabular}[c]{@{}l@{}}Substituting\\ with the most\\ similar word\\ (via word2vec\\ embeddings)
 \end{tabular}} & 10\% & 0.83 & 0.84 & 0.81 & 0.88 & 0.79 & 5.63\\
 & 20\% & 0.81 & 0.82 & 0.78 & 0.88 & 0.75 & 6.23\\
 & 30\% & 0.81 & 0.82 & 0.80 & 0.83 & 0.79 & 6.23\\
 & 40\% & 0.81 & 0.82 & 0.78 & 0.88 & 0.75 & 6.32\\
 & 50\% & 0.81 & 0.80 & 0.86 & 0.75 & 0.88 & 6.20\\
 & 60\% & 0.83 & 0.84 & 0.81 & 0.88 & 0.79 & 6.10\\
 & 70\% & 0.71 & 0.70 & 0.73 & 0.67 & 0.75 & 6.32\\
 & 80\% & 0.75 & 0.75 & 0.75 & 0.75 & 0.75 & 6.35\\
 & 90\% & 0.77 & 0.78 & 0.76 & 0.79 & 0.75 & 6.26\\
 \hdashline
\multirow{9}{*}{\begin{tabular}[c]{@{}l@{}}Substituting\\ synonyms\\ (via\\ WordNet)\end{tabular}} & 10\% & 0.77 & 0.79 & 0.72 & 0.88 & 0.67 & 3.55\\
 & 20\% & 0.75 & 0.75 & 0.75 & 0.75 & 0.75 & 3.97\\
 & 30\% & 0.77 & 0.78 & 0.76 & 0.79 & 0.75 & 4.12\\
 & 40\% & 0.75 & 0.77 & 0.71 & 0.83 & 0.67 & 4.01\\
 & 50\% & 0.75 & 0.75 & 0.75 & 0.75 & 0.75 & 3.86\\
 & 60\% & 0.71 & 0.70 & 0.73 & 0.67 & 0.75 & 4.33\\
 & 70\% & 0.81 & 0.82 & 0.80 & 0.83 & 0.79 & 3.98\\
 & 80\% & 0.81 & 0.82 & 0.78 & 0.88 & 0.75 & 4.08\\
 & 90\% & 0.77 & 0.78 & 0.74 & 0.83 & 0.71 & 4.16
\end{tabular}
\end{adjustbox}
\caption{Performance of the fine-tuned BERT model and similarity between original and perturbed texts (Wasserstein distance $W_1$).}
\label{tab:perf}
\end{table}

\begin{figure*}[t!]
\includegraphics[width=0.9\linewidth]{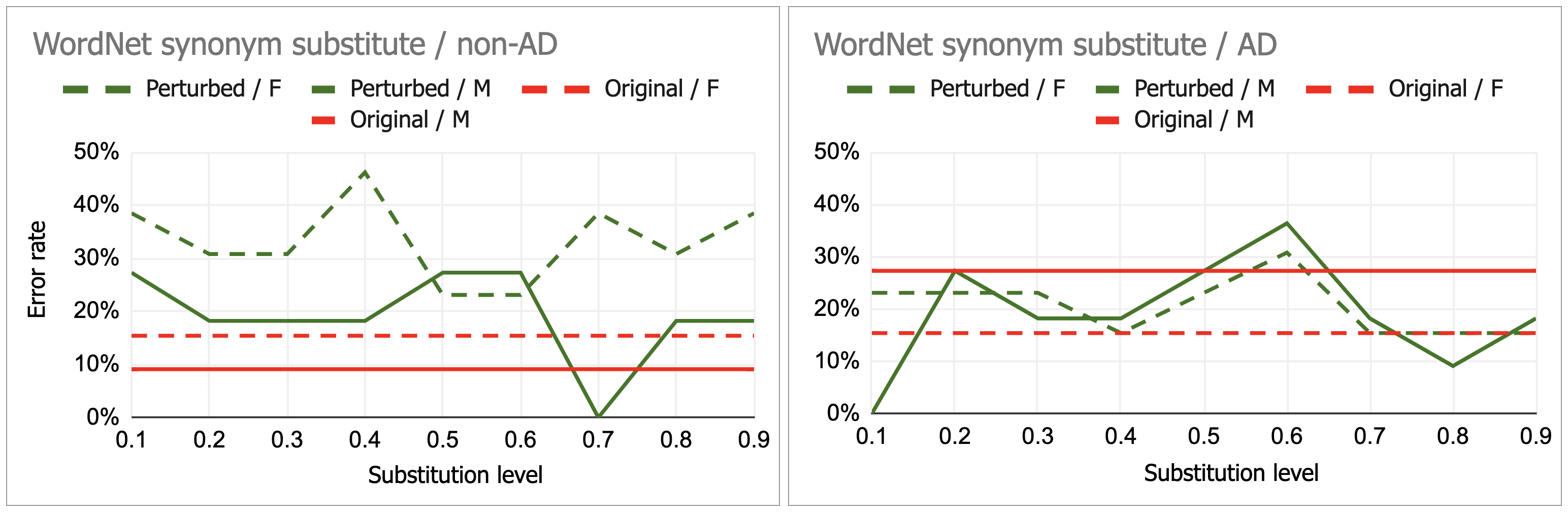}
\centering
\caption{Differences in error rate between genders, by class. Here, M means 'male' and F means 'female'.}
\label{fig:synonym-sex}
\end{figure*}

\section{Discussion}

\subsection{Change in Classification Performance}

\textbf{Undesired change:} As we have mentioned in Section~\ref{sec:alterations}, some types of text alterations, such as back translation, synonym substitution and embedding-based substitution, represent natural noise that can occur in user-generated texts. Changes in classification performance are not desired in this case because we want the model to be robust towards multiple paraphrases and use of synonyms. Our fine-tuned BERT model behaves in a robust way in terms of F1 and accuracy scores when up to 40\% of words are substituted with similar words based on word2vec embeddings (the F1 score decreases by 1\% and accuracy - by 2\%, both changes not significant with McNemar's test $p>$0.65) or synonyms (a decrease in 4-8\% in F1 and accuracy, both changes not significant with McNemar's test $p>$0.15).

Interestingly, recall and specificity values seem to show the opposite results here - specificity decreases by up to 21\%, while recall stays on the same level or even increases by 4-9\%. Substituting words with their similar alternatives or synonyms may be understood as increasing the level of lexical complexity, i.e. the model is introduced with multiple, maybe less usual, ways to express the same meaning. It is known that lexical complexity is one of the prominent ways that allow detecting cognitive impairment from language. Thus such an implicit way to change the lexical complexity of texts seems to help the BERT model in reducing the amount of true positive errors while detecting AD. However, more than 40\% of such substitutions may make the original texts less realistic, which, as we see from the results, substantially reduces model performance, including reducing recall level.

\begin{table}[!t]
\begin{adjustbox}{max width=1\linewidth, center}
\begin{tabular}{llrrrrrr}
\multirow{2}{*}{\textbf{\begin{tabular}[c]{@{}l@{}}Type of\\ perturbation\end{tabular}}} & \multicolumn{5}{c}{\textbf{Correlation between $W_1$ and}} \\
 & \multicolumn{1}{l}{\textbf{Acc}} & \multicolumn{1}{l}{\textbf{F1}} & \multicolumn{1}{l}{\textbf{Prec}} & \multicolumn{1}{l}{\textbf{Rec}} & \multicolumn{1}{l}{\textbf{Spec}} \\
 \hline \hline
\begin{tabular}[c]{@{}l@{}}Deleting\\ informational\\ units\end{tabular} & 0.24 & 0.56 & -0.07 & 0.52 & -0.21 \\
\hline
\begin{tabular}[c]{@{}l@{}}Substituting\\ with the most\\ similar word\\ (via word2vec\\ embeddings)\end{tabular} & 0.77 & 0.89 & 0.89 & 0.87 & -0.91 \\
\hline
\begin{tabular}[c]{@{}l@{}}Substituting\\ synonyms\\ (via\\ WordNet)\end{tabular} & -0.26 & -0.37 & 0.10 & -0.45 & 0.41
\end{tabular}
\end{adjustbox}
\caption{Correlation between similarity and performance metrics.}
\label{tab:wasserstein}
\end{table}

\textbf{Desired change:} Other types of text alterations, such as removal of information units or tokens representing filled pauses, are not considered to be natural noise. As these characteristics of language are clinically important in detecting cognitive impairment, the models should be sensitive to such changes in language. Our results show that the fine-tuned BERT model ignores completely removal of filled pauses. Performance of the model decreases by 3-5\% of F1 as a reaction to deleting different types of information units but this change is not significant (McNemar's test $p>$0.18). This change in performance is similar to the change caused by synonym substitution and shows that the model is not sensitive enough to removal of clinically relevant information. 

This leads us to inspect how each type of alterations affects distributional shift from the original text and whether there is a relation between the shift and model's performance.

\subsection{Correlation with Distributional Shift}

Inspired by \cite{lee2018simple} where hidden activations were used to detect out-of-distribution samples for images and by \citet{rychener2020sentence} applied this method to text, we used sentence embeddings produced by BERT to quantify the distributional shift among the original test set and its perturbed versions. To understand the level of dissimilarity among the versions of test sets, we calculated the 1-Wasserstein distance (``earth mover distance", $W_1$), since it measures the minimum cost to turn one probability distribution into another (Table~\ref{tab:perf}).

W$_1$ values show that deletion of information units and filled pauses has the lowest effect on the original text, while word2vec-based substitution shifts the distribution further away from the original. Correlation between performance metrics and W$_1$ is not consistent across different types of text alterations (Table~\ref{tab:wasserstein}): it is strongly positive between F1 and accuracy scores in case of embedding-based substitutions (0.77 and 0.89), positive but less strong (0.24 and 0.56) in case of deleting clinically relevant information, and negative in case of synonym substitution (-0.26 and -0.37). These inconsistencies imply that the lack of sensitivity in BERT models is caused by intrinsic model reasons rather than distributional shift of test data.

\vspace{-0.2em}
\subsection{Differences Based on Gender}

In order to understand if BERT performance is biased towards any gender, we analyse the rate of error within each gender group and how the error rate is changing with additional amount of text alterations. The results of this analysis do not reveal any differences between males and females within the class of AD data samples. However when it comes to the non-AD class, BERT tends to misclassify the text samples produced by female subjects significantly more often than those produced by males, across all types of text alterations. The effect is pronounced the most in the case of synonym substitution (see Figure~\ref{fig:synonym-sex}), where the error rate of classifying female-produced samples is 14\% higher on average than that of male-produced samples\footnote{Also significantly different based on t-test, $p<$0.005.}. Given that both training and test sets of the dataset are well balanced, such a difference implies the pre-trained BERT model is gender-biased and this bias is not eliminated during fine-tuning. 

\vspace{-0.2em}
\section{Limitations and Conclusions}
\label{sec:conclusions}
\vspace{-0.3em}
In this work, we analysed how the controlled amount of desired and undesired text alterations impacts BERT classification performance in the domain of AD detection. We showed that BERT is robust enough to the natural linguistic noise, although the model is biased towards text samples of non-AD females. On the other hand, BERT is not sensitive enough to removal of clinically relevant information. This lack of sensitivity is not directly influenced by distributional shift.

This work is a first step towards investigating BERT models' robustness and sensitivity in the domain of AD detection from text, and we only report empirical results of one BERT model fine-tuned and tested on one dataset. More work should be done in this area to ensure the results are widely generalizable within the domain. Textual data used in our experiment represent transcribed conversational speech and as such, may be quite different from other types of texts, e.g. written text. Future work is necessary to see if the effect of text alterations remain the same with other types of text.

\bibliography{anthology,custom}
\bibliographystyle{acl_natbib}

\newpage
\appendix



\end{document}